%% file: lrec2026-example.tex
\documentclass[10pt, a4paper]{article}
\usepackage{hyperref}
\usepackage[most]{tcolorbox}
\usepackage{cleveref}
\usepackage{graphicx}
\usepackage{array}     
\usepackage{multirow}  
\usepackage{geometry} 
\usepackage{setspace}
\usepackage{subcaption}
\usepackage{booktabs,tabularx}
\usepackage{mathrsfs}
\usepackage{xcolor}
\usepackage[table]{xcolor}
\usepackage{enumitem}
\usepackage{makecell}
\crefformat{section}{\S#2#1#3}
\crefformat{subsection}{\S#2#1#3}
\crefformat{subsubsection}{\S#2#1#3}
\crefrangeformat{section}{\S\S#3#1#4 to~#5#2#6}
\crefmultiformat{section}{\S\S#2#1#3}{ and~#2#1#3}{, #2#1#3}{ and~#2#1#3}
\usepackage{lrec2026} 
\usepackage{colortbl}

\newtcolorbox{resultbox}[1][]{%
    colback=black!3,
    colframe=black!3,
    notitle,
    sharp corners,
    borderline west={2pt}{0pt}{gray!80!black},
    enhanced,
    breakable,
    boxsep=0pt,
    left=4pt,right=2pt,top=2pt,bottom=2pt,
    }

\title{\texttt{MUNIChus}: Multilingual News Image Captioning Benchmark}

\name{Yuji Chen$^{1}$, Alistair Plum$^{2}$, Hansi Hettiarachchi$^{1}$, Diptesh Kanojia$^{3}$, \\
      {\bfseries \large Saroj Basnet$^{4}$, Marcos Zampieri$^{4}$ and Tharindu Ranasinghe$^{1}$}}

\address{$^{1}$Lancaster University, UK
         $^{2}$University of Luxembourg, Luxembourg \\
         $^{3}$University of Surrey, UK
         $^{4}$George Mason University, USA \\
         t.ranasinghe@lancaster.ac.uk}

\abstract{
 The goal of news image captioning is to generate captions by integrating news article content with corresponding images, highlighting the relationship between textual context and visual elements. The majority of research on news image captioning focuses on English, primarily because datasets in other languages are scarce. To address this limitation, we \textbf{create the first multilingual news image captioning benchmark, \texttt{MUNIChus}}, comprising 9 languages, including several low-resource languages such as Sinhala and Urdu. We evaluate various state-of-the-art neural news image captioning models on \texttt{MUNIChus} and find that news image captioning remains challenging. We also make \texttt{MUNIChus} publicly available with over 20 models already benchmarked. \texttt{MUNIChus} opens new avenues for further advancements in developing and evaluating multilingual news image captioning models. 
 \\ \newline \Keywords{News Image Captioning, Multimodal Large Language Models, Text Generation} }

\begin{document}

\maketitleabstract

\section{Introduction}

Image captioning is a task that bridges language and vision~\cite{liu-etal-2021-visual}, where the goal is to generate a semantically accurate description of the visual content in a syntactically correct manner~\cite{9706348, XU2023126287}. Image captioning can have profound impacts, particularly in assisting visually impaired individuals by enabling a better understanding of the content within an image~\cite{DANESHFAR2024109288}. Therefore, the task has attracted significant interest and achieved notable advancements in recent years~\cite{10.1145/3617592}. The recent introduction of multimodal large language models (MLLMs)~\cite{10.1145/3672758.3672824, caffagni-etal-2024-revolution, XIAO2025102888} has fuelled the development of image captioning models and provided state-of-the-art results~\cite{agarwal2024methods, 9706348}. Furthermore, MLLMs have also been leveraged for multilingual image captioning, expanding their applicability across diverse linguistic contexts~\cite{ramos-etal-2023-lmcap, 10.24963/ijcai.2024/887}.

News image captioning, a variant of the image captioning task, requires generating an informative caption, rich in entities such as the names of people or news events, based on the provided news image and associated news article~\cite{rajakumar-kalarani-etal-2023-lets}. Typically, a news image illustrates a portion of the article, with the caption linking the image content to the article. Ideally, readers should grasp the article's essence by simply browsing its images and their captions~\cite{qu-etal-2024-visually}.

\begin{figure}
    \centering
    
    \begin{subfigure}{\columnwidth}
        \captionsetup{labelformat=empty} 
        \centering
        \begin{minipage}{0.5\columnwidth}
            \centering
            \includegraphics[width=\textwidth]{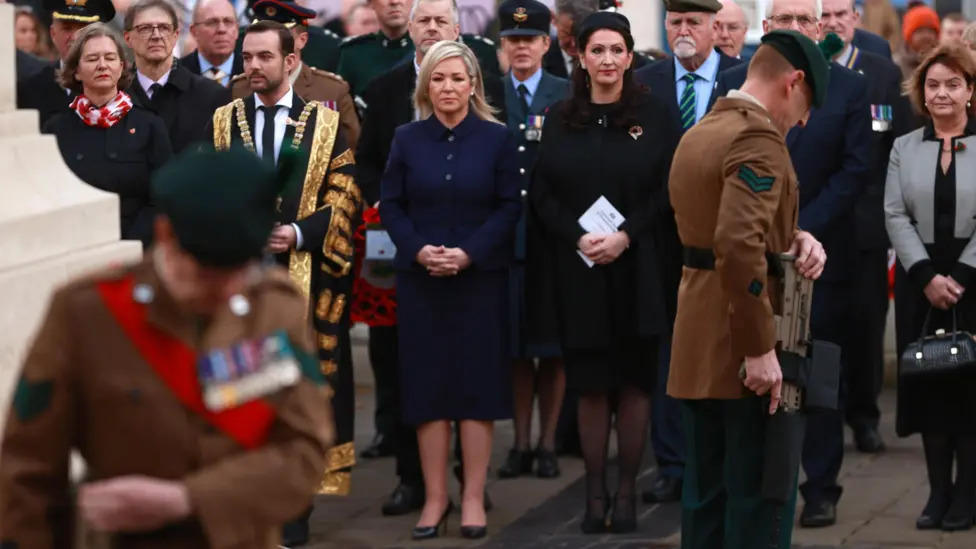}
        \end{minipage}
        \hfill
        \begin{minipage}{0.45\columnwidth}
            \raggedright
            \setstretch{0.5}
            {\scriptsize \textbf{News Image Caption:} \\ Michelle O'Neill attended the Belfast ceremony alongside Deputy First Minister Emma Little-Pengelly of the Democratic Unionist Party (DUP).}

            \noindent\rule{0.9\columnwidth}{0.4pt}

            {\scriptsize \textbf{Generic Image Caption:} \\ A crowd of people standing around each other.}
        \end{minipage}
        \caption{} 
        \label{fig:political_event}
    \end{subfigure}
    
    \begin{subfigure}{\columnwidth}
        \captionsetup{labelformat=empty} 
        \centering
        \begin{minipage}{0.5\columnwidth}
            \centering
            \includegraphics[width=\textwidth]{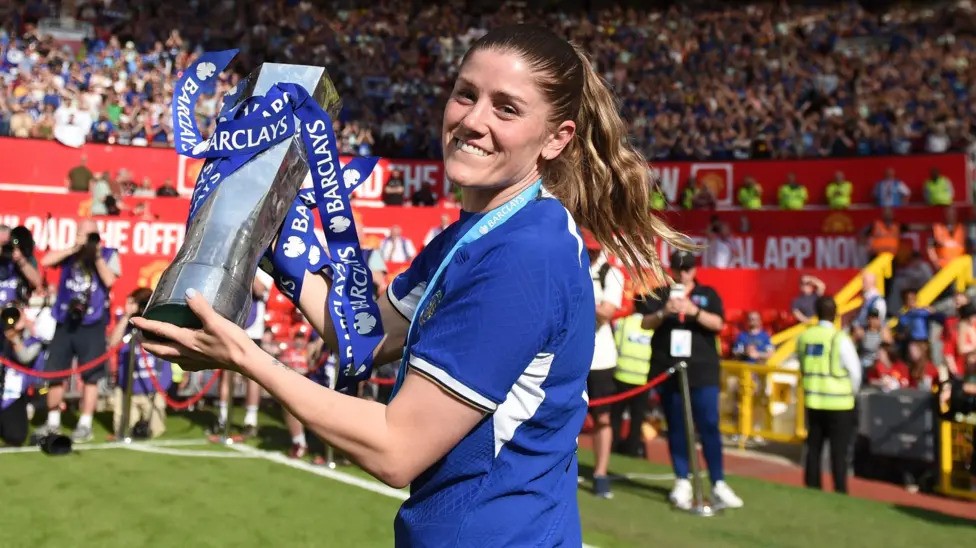}
        \end{minipage}
        \hfill
        \begin{minipage}{0.45\columnwidth}
            \raggedright
            \setstretch{0.5}
            {\scriptsize \textbf{News Image Caption:} \\ Maren Mjelde won the Women's Super League in her final season with Chelsea.}

            \noindent\rule{0.9\columnwidth}{0.4pt}

            {\scriptsize \textbf{Generic Image Caption:} \\ A woman holding a trophy in front of a crowd.}
        \end{minipage}
        \caption{} 
        \label{fig:sports_event}
    \end{subfigure}
    
    \captionsetup{width=\columnwidth}
    \vspace{-0.8cm}
    \caption{Comparison of news and generic image captions for two different images. The generic image captions were generated by BLIP~\cite{pmlr-v162-li22n}.}
\end{figure}

Generic image captions are different to news image captions as they are descriptive rather than interpretative, and referents to objects are generic rather than specific~\cite{liu-etal-2021-visual}. For example, while a caption such as “A crowd of people standing around each other” properly describes the image in Figure~\ref{fig:political_event}, it fails to capture the underlying context that is taking place in this picture, such as “Why are the people standing together, and who are they?”. Similarly, in Figure~\ref{fig:sports_event}, the generic caption “A woman holding a trophy in front of a crowd” offers a basic description but omits the specific significance conveyed by the news caption “Maren Mjelde won the Women's Super League in her final season with Chelsea,” such as the identity of the woman and the event’s importance. Consequently, news image captioning models adopt distinct approaches compared to generic image captioning models, often integrating news content as an input to better contextualise the visual information.

\begin{table*}[t]
    \centering
    \resizebox{\linewidth}{!}{
    \begin{tabular}{lcccccccccc}
        \toprule
         & {\color{yellow}\rule{1mm}{3mm}} \textbf{Arabic} & {\color{green}\rule{1mm}{3mm}} \textbf{Chinese} & {\color{green}\rule{1mm}{3mm}} \textbf{English} & {\color{green}\rule{1mm}{3mm}} \textbf{French} & {\color{yellow}\rule{1mm}{3mm}} \textbf{Hindi} & {\color{red}\rule{1mm}{3mm}} \textbf{Indonesian} & {\color{yellow}\rule{1mm}{3mm}} \textbf{Japanese} & {\color{red}\rule{1mm}{3mm}} \textbf{Sinhala} & {\color{red}\rule{1mm}{3mm}} \textbf{Urdu} & \textbf{All} \\
         \midrule
        \textbf{Family} & \makecell{Afro-Asiatic\\(Semitic)} & Sino-Tibetan$^*$ & \makecell{Indo-European\\(Germanic)} & \makecell{Indo-European\\(Romance)} & \makecell{Indo-European\\(Indo-Aryan)} & Austronesian & Japonic$^\dagger$ & \makecell{Indo-European\\(Indo-Aryan)} & \makecell{Indo-European\\(Indo-Aryan)} & --- \\
        \textbf{$|$Train$|$} & 5,119 & 9,389 & 79,195 & 10,247 & 12,566 & 12,137 & 7,641 & 2,418 & 6,602 & 145,314 \\
        \textbf{$|$Test$|$} & 999 & 999 & 1,000 & 999 & 1,000 & 1,000 & 1,000 & 998 & 998 & 8,993 \\
        \textbf{Unique Articles} & 2,289 & 2,922 & 38,558 & 2,853 & 2,760 & 1,952 & 3,805 & 1,046 & 2,478 & 58,663 \\
        \textbf{Avg Images/Article} & 2.67 & 3.56 & 2.08 & 3.94 & 4.92 & 6.73 & 2.27 & 3.27 & 3.07 & 3.61 \\
        \textbf{Avg Content Tokens} & 1,010 & 1,519 & 461 & 1,510 & 1,968 & 1,794 & 1,287 & 1,194 & 1,792 & 1,412 \\
        \textbf{Avg Caption Tokens} & 12.70 & 17.28 & 13.66 & 17.23 & 12.32 & 16.24 & 20.63 & 16.39 & 15.68 & 15.98 \\
        \textbf{Avg Title Tokens} & 10.85 & 14.38 & 7.76 & 14.23 & 14.03 & 14.35 & 18.12 & 10.16 & 18.28 & 13.68 \\
        \bottomrule
    \end{tabular}
    }
   \caption{Dataset statistics for \texttt{MUNIChus} across nine languages showing language family, number of images in train/test splits, unique articles, and average token counts for images per article, content, captions, and titles. Languages are categorized as high-resource {\color{green}\rule{1mm}{3mm}}, mid-resource {\color{yellow}\rule{1mm}{3mm}}, and low-resource {\color{red}\rule{1mm}{3mm}} following \protect~\citet{joshi-etal-2020-state}. $^*$Chinese tokenised using Jieba. $^\dagger$Japanese tokenised using MeCab.}
    \label{tab:tasks_dataset}
\end{table*}

Over the years, researchers have introduced several benchmark datasets, such as \textsc{Visual News}~\citelanguageresource{liu-etal-2021-visual}, \textsc{NYTimes800k}~\citelanguageresource{9156456}, and \textsc{GoodNews}~\citelanguageresource{8953244} to perform news image captioning. However, these datasets are exclusively focused on English, limiting the scope of existing news image captioning models, which have been trained and evaluated solely in English, hindering progress in adapting such models to other languages. While techniques such as multilingual~\cite{ramos-etal-2023-lmcap, kim-etal-2023-pr, 10.1145/3582649.3582658} and cross-lingual learning~\cite{wu-etal-2023-cross2stra, 10.1145/3444685.3446322, 10.1145/3634917} have shown promise in improving generic image captioning, particularly for low-resource languages, they remain largely unexplored in the news domain due to the absence of multilingual datasets.

To address this gap, our work introduces \textbf{\texttt{MUNIChus}}, the first multilingual news image captioning benchmark. \texttt{MUNIChus} comprises more than $700,000$ news images, each accompanied by the news article, article headline and a corresponding caption. \texttt{MUNIChus} spans over $9$ languages, including several low-resource languages such as Sinhala~\cite{de2019survey, hettiarachchi-etal-2024-nsina, ranasinghe-etal-2025-sinhala} and Urdu . The \textbf{main contributions} of this work can be summarised as:

\begin{enumerate}
    \item We \textbf{release \texttt{MUNIChus}}, the first multilingual news image captioning benchmark\footnote{The dataset is available in HuggingFace following \url{https://huggingface.co/datasets/tharindu/MUNIChus}}. This is the largest publicly available news image captioning dataset, covering over 700,000 news images across diverse language families.  \texttt{MUNIChus} compromises of 9 languages including three low-resource languages according to~\citet{joshi-etal-2020-state}. 
    \item We \textbf{evaluate} more than 20 models on \texttt{MUNIChus}, including state-of-the-art MLLMs and generic image captioning models. We \textbf{show} that multilingual news image captioning remains a challenging task despite advances in MLLMs, even in high-resource languages. 
\end{enumerate}

\section{\texttt{MUNIChus}: Multilingual News Image Captionning Benchmark}
We first present the data collection methodology we used, followed by a detailed statistical analysis of \texttt{MUNIChus}.

\begin{figure*}[t]
\centering
\begin{tcolorbox}[
    colback=gray!5,
    colframe=gray!75,
    fonttitle=\bfseries,
    width=0.95\textwidth,
    boxrule=0.5pt
]
\small
\texttt{You are writing a caption for a newspaper image.}

\vspace{0.3em}
\texttt{Given the image and this news article excerpt:}\\
\texttt{[News content]}

\vspace{0.3em}
\texttt{Task: Write a concise, informative caption for this image in \{language\}.}

\vspace{0.3em}
\texttt{Guidelines:}

\texttt{- Write in \{language\} language only}

\texttt{- Keep it brief}

\texttt{- Identify and include: people's names, locations, and organisations}

\texttt{- Connect what you see in the image to the news context}

\texttt{- Use journalistic style (factual, clear, objective)}

\texttt{- Focus on the main subject of the image}

\vspace{0.3em}
\texttt{Caption in \{language\}:}
\end{tcolorbox}
\caption{Prompt template used for zero-shot image captioning. The \{language\} placeholder is replaced with the target language name (e.g., ``English'', ``Arabic'') at inference time.}
\label{fig:prompt}
\end{figure*}

\texttt{MUNIChus} comprises images, corresponding news articles, captions, news article headlines and associated metadata obtained from the British Broadcasting Corporation (BBC). We selected BBC as the data source due to its rigorous editorial charter. Moreover, the BBC’s extensive international presence ensures comprehensive coverage across a wide range of topics and languages. Owing to these qualities, BBC news content has also been widely utilised in the construction of several NLP datasets such as \texttt{XL-Sum}~\cite{hasan-etal-2021-xl}.

We developed a Python-based scraper to collect BBC news articles in the languages listed in Table~\ref{tab:tasks_dataset}, published before December 31, 2024. To ensure data quality, we first removed images with a height or width of less than $180$ pixels. Next, we retained only those examples whose captions contain more than three words. While our experiments utilise only images and articles, the \texttt{MUNIChus} also provides additional metadata, such as article titles, which can be used in future studies.

As shown in Table~\ref{tab:tasks_dataset}, the final dataset consisted of 145,314 training images and 8,993 test images across nine languages, sourced from 58,663 unique BBC news articles. English contributes the most to the corpus, with over $79,000$ training images and $38,000$ unique articles. The average content length varies widely across languages, ranging from $461$ tokens in English to $1,968$ in Hindi. Furthermore, the average number of images per article also shows considerable variation, with Indonesian having the highest density at $6.73$ images per article, while English has the lowest at $2.08$ images per article. The corpus exhibits diversity in caption and title lengths as well, with Japanese having the longest captions ($20.63$ tokens on average) and titles ($18.12$ tokens on average), while English has the shortest titles ($7.76$ tokens on average).

\subsection{Evaluation}
We employed BLEU-4~\cite{papineni-etal-2002-bleu} and CIDEr~\cite{oliveira-dos-santos-etal-2021-cider} as the primary evaluation metrics for assessing models on \texttt{MUNIChus}, comparing each generated caption against its corresponding reference caption. These metrics have been widely adopted in previous news image captioning studies~\cite{liu-etal-2021-visual, qu-etal-2024-visually}. Although more advanced metrics such as BERTScore~\cite{Zhang*2020BERTScore} and BLEURT~\cite{sellam-etal-2020-bleurt} have been proposed for text generation tasks, they lack support for Sinhala and Urdu. Since our goal was to provide a comprehensive evaluation across all languages in \texttt{MUNIChus}, we restricted our evaluation to BLEU-4 and CIDEr. Additionally, prior work~\cite{liu-etal-2021-visual, qu-etal-2024-visually} has proposed entity retrieval metrics that assess the overlap of named entities between generated and reference captions. While this approach offers valuable insights into factual consistency, its implementation requires robust named entity recognition (NER) models. Given the limited availability and suboptimal performance of NER systems for low-resource languages such as Sinhala and Urdu, particularly within the news domain, we did not incorporate this metric in our evaluation framework.

\section{Methodology}
Following recent advances in NLP, we evaluated several MLLMs for generating news image captions on \texttt{MUNIChus} using two approaches: (i) prompt-based generation (\cref{subsec:prompt}) and (ii) instruction fine-tuning (\cref{subsec:fine}) described below. 

For Chinese and Japanese, we applied language-specific word segmentation before computing BLEU-4~\cite{papineni-etal-2002-bleu} and CIDEr~\cite{oliveira-dos-santos-etal-2021-cider} scores, as these languages lack explicit word boundaries. We used Jieba for Chinese word segmentation, and MeCab for Japanese morphological analysis. Both model predictions and reference captions were tokenised before metric calculation to enable proper word-level n-gram matching.

\subsection{Prompt-based Generation}
\label{subsec:prompt}
We used the following three different prompts to generate the image captions. 

\subsubsection{Zero-shot}
\label{subsec:zero}
In the zero-shot setting, as illustrated in Figure~\ref{fig:prompt}, models received only the task instruction and the news article context without any example captions. The prompt explicitly specified the target language and provided guidelines for generating journalistic captions, including instructions to keep captions concise, identify key entities (people, locations, organisations), maintain factual accuracy, and connect visual content to the news context. This approach tests the model's ability to perform multilingual image captioning solely from its pre-training, without task-specific data.

\subsubsection{Random Few-shot}
\label{subsec:random}
For the random few-shot condition, we augmented each query with three randomly sampled examples from the same language's training set. Each example consisted of an image, its corresponding news article excerpt, and the reference caption. These examples were prepended to the test query to provide the model with in-context learning examples. The random selection ensures that examples are representative of the overall distribution but may not be semantically related to the test instance.

\subsubsection{Similar Few-shot}
\label{subsec:similar}
In the similar few-shot setting, we selected three examples that were most semantically similar to the test instance. We computed similarity by encoding images using the \texttt{nomic-embed-vision-v1.5} vision encoder~\cite{nussbaum2024nomic} and retrieving the three training instances with the highest cosine similarity to the test image embedding. This retrieval-augmented approach provides more contextually relevant examples than random selection, potentially enabling better adaptation to the specific characteristics of each test instance through in-context learning, and has been shown to produce better results in many tasks~\cite{xia-etal-2024-rule, mathur-etal-2024-doc, basnet2025evaluating}. The retrieved examples, along with their corresponding news articles and reference captions, were prepended to the test query following the same format as the random few-shot condition.

\subsection{Instruction Fine-tuning}
\label{subsec:fine}

We fine-tuned \texttt{aya-vision-8b} and \texttt{Llama-3.2-11B-Vision-Instruct} using TRL's \texttt{SFTTrainer} under a QLoRA+LoRA setup. For both models, we load the base backbone in 4-bit \texttt{NF4} quantisation with double quantisation and \texttt{bf16} computation, then prepare it for k-bit training before injecting the LoRA adapters. We configure LoRA with rank $r=64$, $\alpha=32$, dropout $=0.1$, and \texttt{bias=none}, and apply it to the standard attention and MLP projection modules (\texttt{q\_proj}, \texttt{k\_proj}, \texttt{v\_proj}, \texttt{o\_proj}, \texttt{gate\_proj}, \texttt{up\_proj}, \texttt{down\_proj}).

For data preparation, we use model-specific collators to construct chat-formatted multimodal training instances. These collators mask user-side prompt tokens and padding tokens with $-100$ so that the training loss is computed only over assistant-side target tokens. Training uses \texttt{bf16} precision with gradient checkpointing, no sequence packing, and a maximum sequence length of 4096 tokens.

We run optimisation for one epoch with a per-device batch size of $1$ and gradient accumulation of $16$. We use fused AdamW with a learning rate of $1.5e-4$, cosine learning-rate decay, a warmup ratio of $0.03$, weight decay of $1e-6$, and gradient clipping at $0.3$. We save checkpoints every $500$ steps (retaining the two most recent checkpoints) and log training metrics every 50 steps. After training, we save the learned LoRA adapters and merge them into the corresponding base model, yielding a standalone checkpoint for deployment.

\subsection{Baselines}
\label{subsec:baseline}
We also employed two general image captioning baselines. We fed the image and generated English captions using BLIP~\cite{pmlr-v162-li22n}, then translated them into the required language using NLLB~\cite{costa2022no}. As the second baseline, we used \texttt{PaliGemma-3b}, a state-of-the-art multilingual image captioning model~\cite{beyer2024paligemma}. 

\section{Results}
\input{result_table}
Table~\ref{tab:results} shows the results for the approaches above when evaluated in the \texttt{MUNIChus} test sets. Overall, instruction fine-tuning substantially outperforms all prompting-based strategies. Fine-tuned \texttt{Llama-3.2-11B} achieves the highest overall BLEU-4 score of 8.40, while fine-tuned \texttt{Aya-vision-8b} achieves the highest overall CIDEr score of 56.34, both representing more than a twofold improvement over the best prompting results. The gains are particularly good for certain languages: Hindi reaches a CIDEr score of 100.12 and Japanese 92.56 under fine-tuned \texttt{Aya-vision-8b}, compared to 91.74 and 21.83, respectively, for the best prompting approach. Fine-tuning also yields the best results across the majority of individual languages in both metrics.

\begin{figure}
    \centering
    
    \begin{subfigure}{\columnwidth}
        \captionsetup{labelformat=empty} 
        \centering
        \begin{minipage}{0.5\columnwidth}
            \centering
            \includegraphics[width=\textwidth]{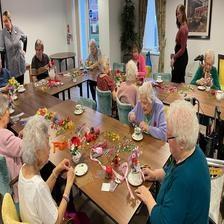}
        \end{minipage}
        \hfill
        \begin{minipage}{0.45\columnwidth}
            \raggedright
            \setstretch{0.5}
            {\scriptsize \textbf{News Image Caption:} \\ Patricia Tomlinson and her friends arranging flowers at Oak Court in Blaby.}

            \noindent\rule{0.9\columnwidth}{0.4pt}

            {\scriptsize \textbf{Generated Caption:} \\ Residents at Oak Court enjoy a flower-arranging class in Blaby.}
        \end{minipage}
        \caption{} 
        \label{fig:examples_error1}
    \end{subfigure}

        \begin{subfigure}{\columnwidth}
        \captionsetup{labelformat=empty} 
        \centering
        \begin{minipage}{0.5\columnwidth}
            \centering
            \includegraphics[width=\textwidth]{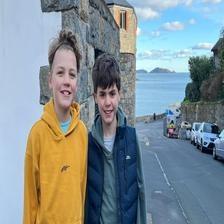}
        \end{minipage}
        \hfill
        \begin{minipage}{0.45\columnwidth}
            \raggedright
            \setstretch{0.5}
            {\scriptsize \textbf{News Image Caption:} \\ Joseph and Thomas are walking because their mum used to be a nurse at the hospice.}

            \noindent\rule{0.9\columnwidth}{0.4pt}

            {\scriptsize \textbf{Generated Caption:} \\ Joseph and Thomas tackle Val De Terres hill for Les Bourgs Hospice fundraiser.}
        \end{minipage}
        \caption{} 
        \label{fig:examples_error2}
    \end{subfigure}

    \captionsetup{width=\columnwidth}
    \vspace{-0.8cm}
    \caption{Comparison of the actual news image caption and the generated caption by the best model - \texttt{GPT-4o} random few-shot approach.}
\end{figure}

Among prompting-based strategies, \texttt{GPT-4o} with random few-shot learning performs best, with an average BLEU-4 score of 3.57 and CIDEr score of 36.17, delivering the highest prompting-based results in five out of nine languages across both metrics. The zero-shot approach using \texttt{Aya-vision-32b} follows as the second-best prompting method, performing best in three languages but showing weaker results in low-resource settings. However, both approaches fall considerably short of the fine-tuned models. For qualitative comparison, Figures~\ref{fig:examples_error1} and~\ref{fig:examples_error2} present several example captions generated by \texttt{GPT-4o} under the random few-shot setting with the corresponding image and reference caption.

The overall results fall within ranges consistent with those observed in previous news image captioning research~\cite{liu-etal-2021-visual}, where BLEU-4 scores typically range from 2 to 4 and CIDEr scores from 20 to 50. However, these modest scores indicate that news image captioning remains a challenging task, particularly given its domain-specific nature.

We describe six key findings (F) next.

\begin{resultbox}
    \textit{{\bf F1:} Traditional Image Captioning Models Demonstrate Dramatic Performance Gap}
\end{resultbox}
Both of our traditional image captioning models performed poorly in multilingual news image captioning, demonstrating that they cannot generate contextual captions for news images. All languages achieved BLEU-4 scores below 0.7 in both baselines, with most below 0.3. The \texttt{BLIP + NLLB} approach averages only 0.20 BLEU-4 overall, while \texttt{Paligemma-3b} performs even worse at 0.03 BLEU-4, including complete failures on languages like Urdu. These results suggest the need for specific models and architectures for news image captioning. 

\begin{figure}
    \centering
    
    \begin{subfigure}{\columnwidth}
        \captionsetup{labelformat=empty} 
        \centering
        \begin{minipage}{0.5\columnwidth}
            \centering
            \includegraphics[width=\textwidth]{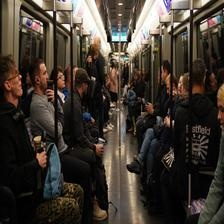}
        \end{minipage}
        \hfill
        \begin{minipage}{0.45\columnwidth}
            \raggedright
            \setstretch{0.5}
            {\scriptsize \textbf{News Image Caption:} \\ The Elizabeth line has opened up more direct journey options across London since it opened in May 2022.}
        \end{minipage}
        \caption{} 
        \label{fig:examples_similar1}
    \end{subfigure}

    \vspace{-8mm}
    \noindent\rule{\columnwidth}{0.4pt}
    \vspace{1mm}

        \begin{subfigure}{\columnwidth}
        \captionsetup{labelformat=empty} 
        \centering
        \begin{minipage}{0.5\columnwidth}
            \centering
            \includegraphics[width=\textwidth]{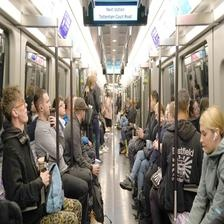}
        \end{minipage}
        \hfill
        \begin{minipage}{0.45\columnwidth}
            \raggedright
            \setstretch{0.5}
            {\scriptsize \textbf{News Image Caption:} \\ TfL says there is an increase in housing growth and employment near stations.}

            \noindent\rule{0.9\columnwidth}{0.4pt}

            {\scriptsize \textbf{Similarity:} \\ 0.973}
        \end{minipage}
        \caption{} 
        \label{fig:examples_similar2}
    \end{subfigure}

    \begin{subfigure}{\columnwidth}
        \captionsetup{labelformat=empty} 
        \centering
        \begin{minipage}{0.5\columnwidth}
            \centering
            \includegraphics[width=\textwidth]{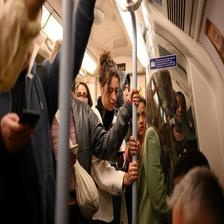}
        \end{minipage}
        \hfill
        \begin{minipage}{0.45\columnwidth}
            \raggedright
            \setstretch{0.5}
            {\scriptsize \textbf{News Image Caption:} \\ Tube services will run as originally planned after industrial action was suspended.}

            \noindent\rule{0.9\columnwidth}{0.4pt}

            {\scriptsize \textbf{Similarity:} \\ 0.953}
        \end{minipage}
        \caption{} 
        \label{fig:examples_similar2}
    \end{subfigure}

    \begin{subfigure}{\columnwidth}
        \captionsetup{labelformat=empty} 
        \centering
        \begin{minipage}{0.5\columnwidth}
            \centering
            \includegraphics[width=\textwidth]{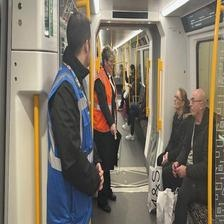}
        \end{minipage}
        \hfill
        \begin{minipage}{0.45\columnwidth}
            \raggedright
            \setstretch{0.5}
            {\scriptsize \textbf{News Image Caption:} \\ The new Metros have linear-style seating and more space for pushchairs and wheelchairs.}

            \noindent\rule{0.9\columnwidth}{0.4pt}

            {\scriptsize \textbf{Similarity:} \\ 0.934}
        \end{minipage}
        \caption{} 
        \label{fig:examples_similar2}
    \end{subfigure}

    \captionsetup{width=\columnwidth}
    \vspace{-0.8cm}
    \caption{Comparison of similar images retrieved for the similar shot approach. The first image is the test instance, and the rest of the images are the most similar images retrieved from the training set. }
    \label{fig:similar1}
\end{figure}

\begin{resultbox}
    \textit{{\bf F2:} Low-Resource Languages Exhibit High Cross-Model Variability}
\end{resultbox}
Low-resource languages like Indonesian, Sinhala, and Urdu exhibit highly inconsistent performance across models in news image captioning, with some architectures handling them reasonably well while others struggle largely. For instance, \texttt{GPT-4o} achieves 5.08 BLEU-4 for Indonesian in random few-shot, while \texttt{Llama-3.2-11B} manages only 0.10 for Sinhala in the same approach, and performance gaps of 5--10x between best and worst models are common for these languages. Instruction fine-tuning reduces this variability to some extent, with both fine-tuned models achieving comparable scores for most languages, though the gap between high-resource and low-resource languages persists. This high variability under prompting suggests that model-specific factors such as training data composition, tokenisation strategies, architectural inductive biases, and the presence of news content in specific languages during pretraining play crucial roles in determining low-resource language performance.

\begin{resultbox}
    \textit{{\bf F3:} Sinhala Shows Consistently Lowest Performance}
\end{resultbox}
Sinhala performs the worst across all models and experimental conditions on \texttt{MUNIChus}, with BLEU-4 scores consistently below 1.2 and CIDEr scores rarely exceeding 17 under prompting-based approaches. Even with instruction fine-tuning, Sinhala remains the weakest language by a large margin, achieving only 0.66--0.74 BLEU-4 and 10.19--11.50 CIDEr, while other languages see dramatic improvements. This substantial and persistent performance gap suggests severe underrepresentation of Sinhala in the pre-training data of the experimented MLLMs, particularly in news-related content, which may include domain-specific vocabulary and cultural references. The consistently poor results across diverse model architectures, prompting strategies, and even after fine-tuning indicate that Sinhala requires targeted intervention, potentially through a dedicated collection of a multimodal news corpus.

\begin{resultbox}
    \textit{{\bf F4:} Model Size Does Not Guarantee Superior Performance}
\end{resultbox}
The relationship between model size and performance on news image captioning appears non-linear and context-dependent. The larger \texttt{Aya-vision-32b} model actually underperforms its smaller 8b counterpart in the majority of the languages, most notably in the random few-shot scenarios. This pattern is further reinforced by the instruction fine-tuning results, where \texttt{Llama-3.2-11B} achieves comparable or superior BLEU-4 scores to fine-tuned \texttt{Aya-vision-8b} despite being among the weakest models under prompting. This finding has important practical implications for deploying vision-language models in this task, as it suggests that smaller, more efficient models may sometimes be preferable to larger alternatives, particularly when task-specific fine-tuning is feasible.

\begin{resultbox}
    \textit{{\bf F5:} No Clear Winner from the Prompting Approaches}
\end{resultbox}
The three prompting strategies yielded mixed results, with no clear indication that any is beneficial for news-image-captioning. Particularly, previous research in MLLMs has shown that a similar few-shot approach improves the results~\cite{ramos-etal-2023-lmcap}. We investigated this issue with the English dataset in \texttt{MUNIChus}. As illustrated in Figure~\ref{fig:similar1}, although the retrieved training images are visually similar to the test instance, they provide limited contextual information relevant to caption generation. These observations indicate that few-shot prompting does not confer substantial benefits in news image captioning.

\begin{resultbox}
    \textit{{\bf F6:} Instruction Fine-tuning Dramatically Outperforms Prompting Strategies}
\end{resultbox}
Instruction fine-tuning yields substantial improvements over all prompting-based approaches, with fine-tuned \texttt{Llama-3.2-11B} and \texttt{Aya-vision-8b} achieving average BLEU-4 scores of 8.40 and 8.37, and CIDEr scores of 44.77 and 56.34, respectively, more than doubling the best prompting results. The gains are consistent across resource levels, with particularly striking improvements in mid-resource languages such as Hindi (100.12 CIDEr) and Japanese (92.56 CIDEr) under fine-tuned \texttt{Aya-vision-8b}. Notably, even relatively small models benefit considerably from fine-tuning: \texttt{Llama-3.2-11B}, which ranked among the weakest models under prompting, becomes highly competitive after fine-tuning, reinforcing finding F4 that model size alone is not the determining factor. However, the improvements for Sinhala remain modest (10.19--11.50 CIDEr), consistent with finding F3, suggesting that fine-tuning alone cannot fully compensate for severe underrepresentation in pre-training data. These results demonstrate that task-specific adaptation through instruction fine-tuning is far more effective than in-context learning for multilingual news image captioning.

Overall, our results suggest that news image captioning is a challenging task even for state-of-the-art MLLMs. Interestingly, current multimodal benchmarks do not contain news image captioning~\cite{li2024seed}. With the release of \texttt{MUNIChus}, we encourage the research community to develop specialised architectures and training strategies that can better leverage contextual information from news articles to generate accurate, entity-rich captions across diverse languages. The persistent challenges observed across all models, particularly for low-resource languages and the limited effectiveness of few-shot prompting, highlight the need for novel approaches that go beyond general-purpose vision-language modelling to address the unique requirements of news image captioning.

\section{Conclusion}

This research presented \texttt{MUNIChus}, the first multilingual news image captioning benchmark, comprising 145,314 training images and 8,993 test images across nine languages. Unlike existing news image captioning datasets that focus primarily on English, \texttt{MUNIChus} addresses the critical gap in multilingual news image captioning by providing diverse linguistic representation, including several low-resource languages such as Sinhala, Indonesian, and Urdu. We evaluated several state-of-the-art multimodal large language models on \texttt{MUNIChus} using prompt-based generation and instruction fine-tuning approaches.

The results show that news image captioning remains a challenging task across all languages. Notably, traditional image captioning models demonstrate dramatic performance gaps, highlighting the domain-specific nature of news image captioning. Furthermore, low-resource languages, particularly Sinhala, exhibit consistently poor performance across all models, suggesting severe underrepresentation in pre-training corpora and highlighting the need for multilingual resources.

In future work, we plan to expand \texttt{MUNIChus} to include additional languages and explore specialised architectures tailored for news image captioning. We also plan to create a comprehensive benchmark suite for multilingual news understanding tasks, incorporating \texttt{MUNIChus} alongside other news-related challenges. We release \texttt{MUNIChus} as an open-access, publicly available dataset alongside trained models\footnote{Fine-tuned \texttt{Aya-vision-8b} model is available at \url{https://huggingface.co/alita9/xl-munichus-CohereLabs-aya-vision-8b} and the fine-tuned \texttt{Llama-3.2-11B} model is available at \url{https://huggingface.co/alita9/xl-munichus-meta-llama-Llama-3.2-11B-Vision-Instruct}} and evaluation scripts to facilitate ongoing research in multilingual multimodal understanding.

\section{Acknowledgement}

Hansi Hettiarachchi is partially supported by the CA21167 COST action UniDive, funded by COST (European Cooperation in Science and Technology).

We acknowledge the EuroHPC Joint Undertaking for awarding us access to Leonardo at CINECA, Italy, to run some LLM experiments. 

Some of the experiments reported in this paper were conducted on the MeluXina high-performance computing infrastructure, an allocation granted by the University of Luxembourg on the EuroHPC supercomputer hosted by LuxProvide.

\section{Ethics Statement}

\texttt{MUNIChus} was collected from publicly available BBC news articles and images, and none of the records were edited in the process. The BBC content is publicly accessible, and our scraping adhered to the site's terms of service. For every instance in \texttt{MUNIChus}, we provide the URL to the original news article, ensuring proper attribution and enabling verification of the source material. We release \texttt{MUNIChus} under the Creative Commons Attribution-NonCommercial-ShareAlike 4.0 International Public License, which prevents users from altering the instances in the dataset and ensures appropriate attribution to the original sources.

The dataset may reflect inherent biases present in news media, including potential geographical, cultural, or topical imbalances in coverage across languages. While we selected the BBC as our data source due to its editorial charter enforcing principles of political impartiality and balanced reporting, we acknowledge that no news source is entirely free from bias. Researchers using \texttt{MUNIChus} should be mindful of these potential biases and consider their implications for downstream applications, particularly when deploying models trained on this dataset in real-world contexts.


\nocite{*}
\section{Bibliographical References}\label{sec:reference}

\bibliographystyle{lrec2026-natbib}
\bibliography{lrec2026-example}

\section{Language Resource References}
\label{lr:ref}
\bibliographystylelanguageresource{lrec2026-natbib}
\bibliographylanguageresource{languageresource}

\end{document}

%% file: result_table.tex
\renewcommand{\arraystretch}{1.2}
\begin{table*}[t]
\begin{center}
\small
\resizebox{\textwidth}{!}{
\begin{tabular}{l || c c c c c c c c c | c || c c c c c c c c c | c} 
\toprule

 & \multicolumn{3}{c}{{\color{green}\rule{1mm}{3mm}} \bf High-resource} & \multicolumn{3}{c}{{\color{yellow}\rule{1mm}{3mm}} \bf Mid-resource} & \multicolumn{3}{c}{{\color{red}\rule{1mm}{3mm}} \bf Low-resource} &  & \multicolumn{3}{c}{{\color{green}\rule{1mm}{3mm}} \bf High-resource} & \multicolumn{3}{c}{{\color{yellow}\rule{1mm}{3mm}} \bf Mid-resource} & \multicolumn{3}{c}{{\color{red}\rule{1mm}{3mm}} \bf Low-resource} &  \\ 
 \cmidrule(r){2-4} \cmidrule(lr){5-7}\cmidrule(lr){8-10} \cmidrule(lr){12-14} \cmidrule(lr){15-17}\cmidrule(l){18-20}
 & \bf En & \bf Fr & \bf Zh & \bf Ar & \bf Hi & \bf Ja & \bf Id & \bf Si & \bf Ur & \bf All & \bf En & \bf Fr & \bf Zh & \bf Ar & \bf Hi & \bf Ja & \bf Id & \bf Si & \bf Ur & \bf All\\
\midrule
\multicolumn{21}{l}{\textbf{\cref{subsec:zero} Zero-shot}} \\
\texttt{Aya-vision-8b} & \cellcolor{teal!24.5} 2.52 & \cellcolor{teal!21.3} 2.19 & \cellcolor{teal!21.3} 2.19 & \cellcolor{teal!14.9} 1.53 & \cellcolor{teal!21.4} 2.20 & \cellcolor{teal!42.7} 4.39 & \cellcolor{teal!26.8} 2.75 & \cellcolor{teal!3.2} 0.33 & \cellcolor{teal!12.6} 1.30 & \cellcolor{teal!21.0} 2.16 & \cellcolor{teal!30.7} 28.13 & \cellcolor{teal!16.0} 14.71 & \cellcolor{teal!19.0} 17.45 & \cellcolor{teal!18.3} 16.78 & \cellcolor{teal!19.9} 18.23 & \cellcolor{teal!26.8} 24.61 & \cellcolor{teal!18.4} 16.89 & \cellcolor{teal!7.3} 6.72 & \cellcolor{teal!14.0} 12.84 & \cellcolor{teal!18.9} 17.37 \\
\texttt{Aya-vision-32b} & \cellcolor{teal!30.4} 3.13 & \cellcolor{teal!29.7} \textbf{3.05} & \cellcolor{teal!24.0} \textbf{2.47} & \cellcolor{teal!19.4} 1.99 & \cellcolor{teal!24.5} 2.52 & \cellcolor{teal!66.4} 6.83 & \cellcolor{teal!29.4} 3.02 & \cellcolor{teal!1.3} 0.13 & \cellcolor{teal!19.0} 1.95 & \cellcolor{teal!27.0} 2.78 & \cellcolor{teal!38.8} 35.55 & \cellcolor{teal!19.8} 18.12 & \cellcolor{teal!27.4} 25.13 & \cellcolor{teal!24.4} 22.42 & \cellcolor{teal!27.4} 25.17 & \cellcolor{teal!50.3} 46.16 & \cellcolor{teal!27.2} 24.94 & \cellcolor{teal!9.8} 9.00 & \cellcolor{teal!22.1} 20.28 & \cellcolor{teal!27.5} 25.20 \\
\texttt{GPT-4o} & \cellcolor{teal!24.7} 2.54 & \cellcolor{teal!16.6} 1.71 & \cellcolor{teal!14.4} 1.48 & \cellcolor{teal!24.5} 2.52 & \cellcolor{teal!34.4} 3.54 & \cellcolor{teal!31.7} 3.26 & \cellcolor{teal!23.0} 2.36 & \cellcolor{teal!11.2} 1.15 & \cellcolor{teal!21.2} 2.18 & \cellcolor{teal!22.5} 2.31 & \cellcolor{teal!43.2} 39.64 & \cellcolor{teal!12.3} 11.31 & \cellcolor{teal!24.3} 22.29 & \cellcolor{teal!31.8} 29.21 & \cellcolor{teal!42.7} 39.13 & \cellcolor{teal!30.7} 28.18 & \cellcolor{teal!26.4} 24.21 & \cellcolor{teal!17.9} 16.42 & \cellcolor{teal!28.6} 26.19 & \cellcolor{teal!28.6} 26.25 \\
\texttt{Llama-3.2-11B} & \cellcolor{teal!23.0} 2.36 & \cellcolor{teal!23.4} 2.41 & \cellcolor{teal!9.1} 0.94 & \cellcolor{teal!10.6} 1.09 & \cellcolor{teal!17.2} 1.77 & \cellcolor{teal!33.2} 3.41 & \cellcolor{teal!21.0} 2.16 & \cellcolor{teal!2.3} 0.24 & \cellcolor{teal!16.0} 1.64 & \cellcolor{teal!17.3} 1.78 & \cellcolor{teal!21.1} 19.37 & \cellcolor{teal!13.5} 12.41 & \cellcolor{teal!9.5} 8.69 & \cellcolor{teal!12.6} 11.59 & \cellcolor{teal!18.7} 17.19 & \cellcolor{teal!18.5} 16.95 & \cellcolor{teal!14.6} 13.39 & \cellcolor{teal!5.0} 4.62 & \cellcolor{teal!17.0} 15.61 & \cellcolor{teal!14.5} 13.31 \\
\texttt{Phi-3.5-vision} & \cellcolor{teal!22.1} 2.27 & \cellcolor{teal!21.5} 2.21 & \cellcolor{teal!18.3} 1.88 & \cellcolor{teal!17.8} 1.83 & \cellcolor{teal!25.6} 2.63 & \cellcolor{teal!39.2} 4.03 & \cellcolor{teal!24.8} 2.55 & \cellcolor{teal!4.8} 0.49 & \cellcolor{teal!19.2} 1.97 & \cellcolor{teal!21.5} 2.21 & \cellcolor{teal!27.5} 25.21 & \cellcolor{teal!15.2} 13.92 & \cellcolor{teal!21.8} 19.98 & \cellcolor{teal!20.5} 18.76 & \cellcolor{teal!24.3} 22.29 & \cellcolor{teal!29.4} 26.94 & \cellcolor{teal!20.1} 18.45 & \cellcolor{teal!8.9} 8.15 & \cellcolor{teal!18.7} 17.12 & \cellcolor{teal!20.5} 18.87 \\
\texttt{Qwen2.5-VL-7B} & \cellcolor{teal!25.5} 2.62 & \cellcolor{teal!22.6} 2.32 & \cellcolor{teal!20.9} 2.15 & \cellcolor{teal!19.3} 1.98 & \cellcolor{teal!28.0} 2.88 & \cellcolor{teal!55.3} 5.69 & \cellcolor{teal!26.3} 2.70 & \cellcolor{teal!6.0} 0.62 & \cellcolor{teal!23.7} 2.44 & \cellcolor{teal!25.4} 2.61 & \cellcolor{teal!35.6} 32.69 & \cellcolor{teal!15.3} 14.08 & \cellcolor{teal!31.3} 28.68 & \cellcolor{teal!26.4} 24.25 & \cellcolor{teal!31.3} 28.71 & \cellcolor{teal!52.8} 48.46 & \cellcolor{teal!19.7} 18.05 & \cellcolor{teal!13.7} 12.59 & \cellcolor{teal!24.5} 22.46 & \cellcolor{teal!28.0} 25.72 \\
\texttt{Qwen3-VL-8B} & \cellcolor{teal!32.1} 3.30 & \cellcolor{teal!22.4} 2.30 & \cellcolor{teal!22.9} 2.35 & \cellcolor{teal!19.5} 2.00 & \cellcolor{teal!29.1} 2.99 & \cellcolor{teal!44.8} 4.61 & \cellcolor{teal!24.2} 2.49 & \cellcolor{teal!7.6} 0.78 & \cellcolor{teal!23.6} 2.43 & \cellcolor{teal!25.2} 2.59 & \cellcolor{teal!45.5} 41.75 & \cellcolor{teal!17.4} 15.92 & \cellcolor{teal!36.8} \textbf{33.80} & \cellcolor{teal!26.7} 24.46 & \cellcolor{teal!35.1} 32.18 & \cellcolor{teal!55.2} 50.60 & \cellcolor{teal!25.6} 23.46 & \cellcolor{teal!15.9} 14.58 & \cellcolor{teal!30.1} 27.65 & \cellcolor{teal!32.0} 29.38 \\
\midrule
\multicolumn{21}{l}{\textbf{\cref{subsec:random} Random Few-shot}} \\
\texttt{Aya-vision-8b} & \cellcolor{teal!24.6} 2.53 & \cellcolor{teal!18.6} 1.91 & \cellcolor{teal!19.7} 2.02 & \cellcolor{teal!14.6} 1.50 & \cellcolor{teal!20.7} 2.13 & \cellcolor{teal!40.9} 4.20 & \cellcolor{teal!28.5} 2.93 & \cellcolor{teal!5.5} 0.56 & \cellcolor{teal!14.0} 1.44 & \cellcolor{teal!20.4} 2.10 & \cellcolor{teal!30.4} 27.93 & \cellcolor{teal!13.6} 12.51 & \cellcolor{teal!19.5} 17.86 & \cellcolor{teal!17.5} 16.04 & \cellcolor{teal!19.9} 18.23 & \cellcolor{teal!25.9} 23.79 & \cellcolor{teal!19.1} 17.55 & \cellcolor{teal!10.1} 9.25 & \cellcolor{teal!18.5} 16.99 & \cellcolor{teal!18.4} 16.90 \\
\texttt{Aya-vision-32b} & \cellcolor{teal!19.6} 2.02 & \cellcolor{teal!17.9} 1.84 & \cellcolor{teal!15.9} 1.63 & \cellcolor{teal!15.2} 1.56 & \cellcolor{teal!17.9} 1.84 & \cellcolor{teal!32.6} 3.35 & \cellcolor{teal!21.4} 2.20 & \cellcolor{teal!3.3} 0.34 & \cellcolor{teal!16.1} 1.66 & \cellcolor{teal!17.7} 1.82 & \cellcolor{teal!35.7} 32.75 & \cellcolor{teal!14.8} 13.59 & \cellcolor{teal!23.9} 21.92 & \cellcolor{teal!19.8} 18.12 & \cellcolor{teal!22.1} 20.25 & \cellcolor{teal!34.8} 31.92 & \cellcolor{teal!20.7} 19.03 & \cellcolor{teal!10.7} 9.84 & \cellcolor{teal!23.9} 21.92 & \cellcolor{teal!22.9} 21.04 \\
\texttt{GPT-4o} & \cellcolor{teal!37.9} 3.90 & \cellcolor{teal!21.3} 2.19 & \cellcolor{teal!11.7} 1.20 & \cellcolor{teal!24.8} \textbf{2.55} & \cellcolor{teal!100.0} \textbf{10.28} & \cellcolor{teal!29.7} 3.05 & \cellcolor{teal!49.4} 5.08 & \cellcolor{teal!12.1} \textbf{1.24} & \cellcolor{teal!26.1} 2.68 & \cellcolor{teal!34.7} 3.57 & \cellcolor{teal!57.3} 52.59 & \cellcolor{teal!20.6} 18.90 & \cellcolor{teal!18.7} 17.11 & \cellcolor{teal!34.5} 31.61 & \cellcolor{teal!100.0} 91.74 & \cellcolor{teal!23.8} 21.83 & \cellcolor{teal!41.3} 37.92 & \cellcolor{teal!20.8} 19.04 & \cellcolor{teal!35.8} 32.82 & \cellcolor{teal!39.4} 36.17 \\
\texttt{Llama-3.2-11B} & \cellcolor{teal!20.8} 2.14 & \cellcolor{teal!19.1} 1.97 & \cellcolor{teal!4.6} 0.47 & \cellcolor{teal!4.7} 0.48 & \cellcolor{teal!9.8} 1.01 & \cellcolor{teal!19.4} 1.99 & \cellcolor{teal!22.3} 2.29 & \cellcolor{teal!0.9} 0.10 & \cellcolor{teal!6.6} 0.68 & \cellcolor{teal!12.0} 1.23 & \cellcolor{teal!25.5} 23.42 & \cellcolor{teal!13.3} 12.17 & \cellcolor{teal!3.6} 3.27 & \cellcolor{teal!6.7} 6.11 & \cellcolor{teal!11.2} 10.29 & \cellcolor{teal!6.4} 5.90 & \cellcolor{teal!15.2} 13.91 & \cellcolor{teal!2.8} 2.53 & \cellcolor{teal!7.6} 6.97 & \cellcolor{teal!9.6} 8.78 \\
\texttt{Phi-3.5-vision} & \cellcolor{teal!26.8} 2.75 & \cellcolor{teal!20.4} 2.10 & \cellcolor{teal!17.2} 1.77 & \cellcolor{teal!18.9} 1.94 & \cellcolor{teal!27.3} 2.81 & \cellcolor{teal!42.5} 4.37 & \cellcolor{teal!27.6} 2.84 & \cellcolor{teal!6.8} 0.70 & \cellcolor{teal!20.5} 2.11 & \cellcolor{teal!23.1} 2.38 & \cellcolor{teal!31.8} 29.17 & \cellcolor{teal!16.2} 14.85 & \cellcolor{teal!20.7} 18.97 & \cellcolor{teal!22.3} 20.42 & \cellcolor{teal!26.5} 24.31 & \cellcolor{teal!32.1} 29.45 & \cellcolor{teal!21.8} 19.98 & \cellcolor{teal!11.5} 10.52 & \cellcolor{teal!20.9} 19.15 & \cellcolor{teal!22.4} 20.54 \\
\texttt{Qwen2.5-VL-7B} & \cellcolor{teal!28.6} 2.94 & \cellcolor{teal!23.8} 2.45 & \cellcolor{teal!19.4} 1.99 & \cellcolor{teal!18.1} 1.86 & \cellcolor{teal!32.6} 3.35 & \cellcolor{teal!47.7} 4.90 & \cellcolor{teal!25.9} 2.66 & \cellcolor{teal!6.5} 0.67 & \cellcolor{teal!25.2} 2.59 & \cellcolor{teal!26.9} 2.77 & \cellcolor{teal!42.2} 38.76 & \cellcolor{teal!18.5} 16.97 & \cellcolor{teal!27.7} 25.37 & \cellcolor{teal!24.2} 22.21 & \cellcolor{teal!34.9} 32.04 & \cellcolor{teal!47.7} 43.76 & \cellcolor{teal!23.2} 21.29 & \cellcolor{teal!12.1} 11.13 & \cellcolor{teal!26.8} 24.62 & \cellcolor{teal!27.1} 24.86\\
\texttt{Qwen3-VL-8B} & \cellcolor{teal!35.2} 3.62 & \cellcolor{teal!24.1} 2.48 & \cellcolor{teal!13.8} 1.42 & \cellcolor{teal!22.6} 2.32 & \cellcolor{teal!36.4} 3.74 & \cellcolor{teal!48.2} 4.96 & \cellcolor{teal!30.7} 3.16 & \cellcolor{teal!9.3} 0.96 & \cellcolor{teal!27.2} 2.80 & \cellcolor{teal!27.5} 2.83 & \cellcolor{teal!48.7} 44.71 & \cellcolor{teal!19.3} 17.69 & \cellcolor{teal!29.5} 27.05 & \cellcolor{teal!28.9} 26.52 & \cellcolor{teal!38.2} 35.04 & \cellcolor{teal!52.3} 47.98 & \cellcolor{teal!27.8} 25.51 & \cellcolor{teal!16.8} 15.42 & \cellcolor{teal!31.7} 29.08 & \cellcolor{teal!30.3} 27.78 \\

\midrule
\multicolumn{21}{l}{\textbf{\cref{subsec:similar} Similar Few-shot}} \\
\texttt{Aya-vision-8b} & \cellcolor{teal!26.3} 2.70 & \cellcolor{teal!21.1} 2.17 & \cellcolor{teal!18.9} 1.94 & \cellcolor{teal!15.2} 1.56 & \cellcolor{teal!22.4} 2.30 & \cellcolor{teal!40.8} 4.19 & \cellcolor{teal!27.8} 2.86 & \cellcolor{teal!6.2} 0.64 & \cellcolor{teal!16.1} 1.66 & \cellcolor{teal!21.9} 2.25 & \cellcolor{teal!31.2} 28.63 & \cellcolor{teal!15.9} 14.58 & \cellcolor{teal!20.8} 19.08 & \cellcolor{teal!18.1} 16.60 & \cellcolor{teal!21.9} 20.08 & \cellcolor{teal!28.5} 26.14 & \cellcolor{teal!20.9} 19.16 & \cellcolor{teal!10.3} 9.45 & \cellcolor{teal!18.2} 16.69 & \cellcolor{teal!20.6} 18.93 \\
\texttt{Aya-vision-32b} & \cellcolor{teal!32.1} 3.30 & \cellcolor{teal!26.3} 2.70 & \cellcolor{teal!21.7} 2.23 & \cellcolor{teal!22.2} 2.28 & \cellcolor{teal!52.5} 1.92 & \cellcolor{teal!48.9} 5.03 & \cellcolor{teal!46.4} 4.77 & \cellcolor{teal!4.2} 0.43 & \cellcolor{teal!21.1} 2.17 & \cellcolor{teal!34.9} 3.11 & \cellcolor{teal!45.6} 21.86 & \cellcolor{teal!20.0} 18.36 & \cellcolor{teal!27.5} 25.22 & \cellcolor{teal!26.2} 24.05 & \cellcolor{teal!50.4} 46.22 & \cellcolor{teal!41.9} 38.47 & \cellcolor{teal!35.7} 32.75 & \cellcolor{teal!13.2} 12.12 & \cellcolor{teal!25.2} 23.11 & \cellcolor{teal!31.8} 21.13 \\
\texttt{GPT-4o} & \cellcolor{teal!28.7} 2.95 & \cellcolor{teal!18.9} 1.94 & \cellcolor{teal!12.6} 1.30 & \cellcolor{teal!21.9} 2.25 & \cellcolor{teal!45.2} 4.65 & \cellcolor{teal!24.6} 2.53 & \cellcolor{teal!25.7} 2.64 & \cellcolor{teal!8.7} 0.89 & \cellcolor{teal!18.0} 1.85 & \cellcolor{teal!22.7} 2.33 & \cellcolor{teal!50.8} 46.57 & \cellcolor{teal!16.7} 15.33 & \cellcolor{teal!22.1} 20.30 & \cellcolor{teal!27.5} 25.21 & \cellcolor{teal!52.8} 48.38 & \cellcolor{teal!20.8} 19.12 & \cellcolor{teal!25.3} 23.19 & \cellcolor{teal!18.2} 16.68 & \cellcolor{teal!25.3} 23.17 & \cellcolor{teal!28.8} 26.44 \\
\texttt{Llama-3.2-11B} & \cellcolor{teal!21.8} 2.24 & \cellcolor{teal!20.1} 2.07 & \cellcolor{teal!7.8} 0.80 & \cellcolor{teal!11.2} 1.15 & \cellcolor{teal!17.9} 1.84 & \cellcolor{teal!32.8} 3.37 & \cellcolor{teal!23.5} 2.42 & \cellcolor{teal!3.5} 0.36 & \cellcolor{teal!16.8} 1.73 & \cellcolor{teal!17.5} 1.80 & \cellcolor{teal!24.9} 22.84 & \cellcolor{teal!14.5} 13.30 & \cellcolor{teal!9.8} 8.99 & \cellcolor{teal!13.8} 12.65 & \cellcolor{teal!19.8} 18.16 & \cellcolor{teal!19.2} 17.61 & \cellcolor{teal!16.1} 14.76 & \cellcolor{teal!6.8} 6.24 & \cellcolor{teal!18.1} 16.60 & \cellcolor{teal!16.1} 14.79 \\
\texttt{Phi-3.5-vision} & \cellcolor{teal!27.6} 2.84 & \cellcolor{teal!23.2} 2.39 & \cellcolor{teal!18.1} 1.86 & \cellcolor{teal!19.8} 2.04 & \cellcolor{teal!29.2} 3.00 & \cellcolor{teal!43.9} 4.51 & \cellcolor{teal!29.1} 2.99 & \cellcolor{teal!7.8} 0.80 & \cellcolor{teal!22.1} 2.27 & \cellcolor{teal!24.9} 2.56 & \cellcolor{teal!34.1} 31.29 & \cellcolor{teal!17.6} 16.14 & \cellcolor{teal!23.8} 21.82 & \cellcolor{teal!23.9} 21.92 & \cellcolor{teal!28.4} 26.05 & \cellcolor{teal!33.2} 30.45 & \cellcolor{teal!23.5} 21.56 & \cellcolor{teal!13.1} 12.02 & \cellcolor{teal!22.5} 20.63 & \cellcolor{teal!23.7} 21.77 \\
\texttt{Qwen2.5-VL-7B} & \cellcolor{teal!30.2} 3.10 & \cellcolor{teal!25.1} 2.58 & \cellcolor{teal!20.8} 2.14 & \cellcolor{teal!19.4} 1.99 & \cellcolor{teal!35.8} 3.68 & \cellcolor{teal!49.7} 5.11 & \cellcolor{teal!27.9} 2.87 & \cellcolor{teal!8.1} 0.83 & \cellcolor{teal!26.8} 2.76 & \cellcolor{teal!27.9} 2.87 & \cellcolor{teal!43.8} 40.19 & \cellcolor{teal!19.8} 18.16 & \cellcolor{teal!28.9} 26.50 & \cellcolor{teal!26.1} 23.95 & \cellcolor{teal!37.2} 34.13 & \cellcolor{teal!48.5} 44.49 & \cellcolor{teal!25.2} 23.11 & \cellcolor{teal!14.2} 13.02 & \cellcolor{teal!27.8} 25.49 & \cellcolor{teal!28.1} 25.78 \\
\texttt{Qwen3-VL-8B} & \cellcolor{teal!36.4} 3.74 & \cellcolor{teal!25.8} 2.65 & \cellcolor{teal!15.1} 1.55 & \cellcolor{teal!24.2} 2.49 & \cellcolor{teal!39.5} 4.06 & \cellcolor{teal!50.2} 5.16 & \cellcolor{teal!33.8} 3.48 & \cellcolor{teal!10.9} 1.12 & \cellcolor{teal!29.6} 3.04 & \cellcolor{teal!29.9} 3.07 & \cellcolor{teal!50.3} 46.13 & \cellcolor{teal!20.8} 19.08 & \cellcolor{teal!31.5} 28.89 & \cellcolor{teal!29.8} 27.34 & \cellcolor{teal!41.2} 37.79 & \cellcolor{teal!53.8} 49.35 & \cellcolor{teal!29.1} 26.69 & \cellcolor{teal!17.8} 16.32 & \cellcolor{teal!32.8} 30.09 & \cellcolor{teal!32.5} 29.87 \\

\midrule
\multicolumn{21}{l}{\textbf{\cref{subsec:fine} Instruction Fine-tuning}} \\
\texttt{Aya-vision-8b} & \cellcolor{teal!65.1} \textbf{7.40} & \cellcolor{teal!40.1} 2.99 & \cellcolor{teal!50} \textbf{6.68} & \cellcolor{teal!40} 2.61 & \cellcolor{teal!60}8.50 & \cellcolor{teal!80}20.18 & \cellcolor{teal!45}6.23 & \cellcolor{teal!10}0.66 & \cellcolor{teal!15}2.21 & \cellcolor{teal!70.1} 8.37 & \cellcolor{teal!85.1} \textbf{78.27} & \cellcolor{teal!50} \textbf{23.60} & \cellcolor{teal!60}\textbf{67.52} & \cellcolor{teal!40} \textbf{32.66} & \cellcolor{teal!99} \textbf{100.12} & \cellcolor{teal!92} \textbf{92.56} & \cellcolor{teal!45} \textbf{39.30} & \cellcolor{teal!30} 10.19 & \cellcolor{teal!45}\textbf{37.92} & \cellcolor{teal!75.1} \textbf{56.34} \\
\texttt{Llama-3.2-11B} & \cellcolor{teal!70} \textbf{7.40} & \cellcolor{teal!45} \textbf{3.20} & \cellcolor{teal!55} 5.10 & \cellcolor{teal!35} \textbf{2.63} & \cellcolor{teal!55}4.15 & \cellcolor{teal!95} \textbf{21.43} & \cellcolor{teal!45} \textbf{6.57} & \cellcolor{teal!10}0.74 & \cellcolor{teal!35} \textbf{3.73} & \cellcolor{teal!80} \textbf{8.40} & \cellcolor{teal!72} 71.59  & \cellcolor{teal!40}22.57 & \cellcolor{teal!45} 39.26 & \cellcolor{teal!35}23.82 & \cellcolor{teal!55}42.40 & \cellcolor{teal!75} 78.21 & \cellcolor{teal!35}29.92 & \cellcolor{teal!25}11.50 & \cellcolor{teal!38}34.83 & \cellcolor{teal!55}44.77 \\
\midrule

\multicolumn{21}{l}{\textbf{\cref{subsec:baseline} Baselines}} \\
\texttt{BLIP + NLLB} & \cellcolor{teal!1.6} 0.16 & \cellcolor{teal!1.2} 0.12 & \cellcolor{teal!0.7} 0.07 & \cellcolor{teal!1.4} 0.14 & \cellcolor{teal!2.5} 0.26 & \cellcolor{teal!5.9} 0.61 & \cellcolor{teal!0.7} 0.07 & \cellcolor{teal!0.4} 0.04 & \cellcolor{teal!4.5} 0.46 & \cellcolor{teal!1.9} 0.20 & \cellcolor{teal!6.4} 5.90 & \cellcolor{teal!3.6} 3.34 & \cellcolor{teal!3.0} 2.78 & \cellcolor{teal!3.6} 3.29 & \cellcolor{teal!5.6} 5.18 & \cellcolor{teal!7.0} 6.41 & \cellcolor{teal!3.8} 3.49 & \cellcolor{teal!1.5} 1.39 & \cellcolor{teal!4.9} 4.50 & \cellcolor{teal!4.4} 4.03 \\
\texttt{Paligemma-3b} & \cellcolor{teal!1.6} 0.17 & \cellcolor{teal!2.0} 0.20 & \cellcolor{teal!1.5} 0.15 & \cellcolor{teal!0.3} 0.03 & \cellcolor{teal!1.4} 0.15 & \cellcolor{teal!5.4} 0.56 & \cellcolor{teal!0.2} 0.02 & \cellcolor{teal!0.1} 0.01 & \cellcolor{teal!0.0} 0.00 & \cellcolor{teal!0.3} 0.03 & \cellcolor{teal!3.8} 3.48 & \cellcolor{teal!2.9} 2.67 & \cellcolor{teal!1.9} 1.75 & \cellcolor{teal!0.7} 0.66 & \cellcolor{teal!1.9} 1.78 & \cellcolor{teal!5.8} 5.33 & \cellcolor{teal!0.4} 0.40 & \cellcolor{teal!0.1} 0.07 & \cellcolor{teal!0.0} 0.00 & \cellcolor{teal!1.4} 1.25 \\
\bottomrule
\end{tabular}
}
\end{center}
\caption{BLEU-4 (left-side) \cite{papineni-etal-2002-bleu} and CIDEr (right-side) \cite{oliveira-dos-santos-etal-2021-cider} evaluation scores across different languages. The best result for each language (any method) is in \textbf{bold}.}
\label{tab:results}
\end{table*}

%% file: lrec2026-example.bbl
\begin{thebibliography}{3}
\expandafter\ifx\csname natexlab\endcsname\relax\def\natexlab#1{#1}\fi

\bibitem[{Biten et~al.(2019)Biten, Gomez, Rusiñol, and Karatzas}]{8953244}
Ali~Furkan Biten, Lluis Gomez, Marçal Rusiñol, and Dimosthenis Karatzas. 2019.
\newblock \href {https://doi.org/10.1109/CVPR.2019.01275} {Good news, everyone! context driven entity-aware captioning for news images}.
\newblock In \emph{2019 IEEE/CVF Conference on Computer Vision and Pattern Recognition (CVPR)}, pages 12458--12467.

\bibitem[{Liu et~al.(2021)Liu, Wang, Wang, and Ordonez}]{liu-etal-2021-visual}
Fuxiao Liu, Yinghan Wang, Tianlu Wang, and Vicente Ordonez. 2021.
\newblock \href {https://doi.org/10.18653/v1/2021.emnlp-main.542} {Visual news: Benchmark and challenges in news image captioning}.
\newblock In \emph{Proceedings of the 2021 Conference on Empirical Methods in Natural Language Processing}, pages 6761--6771, Online and Punta Cana, Dominican Republic. Association for Computational Linguistics.

\bibitem[{Tran et~al.(2020)Tran, Mathews, and Xie}]{9156456}
Alasdair Tran, Alexander Mathews, and Lexing Xie. 2020.
\newblock \href {https://doi.org/10.1109/CVPR42600.2020.01305} {Transform and tell: Entity-aware news image captioning}.
\newblock In \emph{2020 IEEE/CVF Conference on Computer Vision and Pattern Recognition (CVPR)}, pages 13032--13042.

\end{thebibliography}


\begin{thebibliography}{42}
\expandafter\ifx\csname natexlab\endcsname\relax\def\natexlab#1{#1}\fi

\bibitem[{Agarwal and Verma(2024)}]{agarwal2024methods}
Lakshita Agarwal and Bindu Verma. 2024.
\newblock From methods to datasets: A survey on image-caption generators.
\newblock \emph{Multimedia Tools and Applications}, 83(9):28077--28123.

\bibitem[{Basnet et~al.(2025)Basnet, Farabi, Ranasinghe, Kanojia, and Zampieri}]{basnet2025evaluating}
Saroj Basnet, Shafkat Farabi, Tharindu Ranasinghe, Diptesh Kanojia, and Marcos Zampieri. 2025.
\newblock Evaluating open-source vision-language models for multimodal sarcasm detection.
\newblock \emph{arXiv preprint arXiv:2510.11852}.

\bibitem[{Beyer et~al.(2024)Beyer, Steiner, Pinto, Kolesnikov, Wang, Salz, Neumann, Alabdulmohsin, Tschannen, Bugliarello, Unterthiner, Keysers, Koppula, Liu, Grycner, Gritsenko, Houlsby, Kumar, Rong, Eisenschlos, Kabra, Bauer, Bošnjak, Chen, Minderer, Voigtlaender, Bica, Balazevic, Puigcerver, Papalampidi, Henaff, Xiong, Soricut, Harmsen, and Zhai}]{beyer2024paligemma}
Lucas Beyer, Andreas Steiner, André~Susano Pinto, Alexander Kolesnikov, Xiao Wang, Daniel Salz, Maxim Neumann, Ibrahim Alabdulmohsin, Michael Tschannen, Emanuele Bugliarello, Thomas Unterthiner, Daniel Keysers, Skanda Koppula, Fangyu Liu, Adam Grycner, Alexey Gritsenko, Neil Houlsby, Manoj Kumar, Keran Rong, Julian Eisenschlos, Rishabh Kabra, Matthias Bauer, Matko Bošnjak, Xi~Chen, Matthias Minderer, Paul Voigtlaender, Ioana Bica, Ivana Balazevic, Joan Puigcerver, Pinelopi Papalampidi, Olivier Henaff, Xi~Xiong, Radu Soricut, Jeremiah Harmsen, and Xiaohua Zhai. 2024.
\newblock Paligemma: A versatile 3b vlm for transfer.
\newblock \emph{arXiv preprint arXiv:2407.07726}.

\bibitem[{Biten et~al.(2019)Biten, Gomez, Rusiñol, and Karatzas}]{8953244}
Ali~Furkan Biten, Lluis Gomez, Marçal Rusiñol, and Dimosthenis Karatzas. 2019.
\newblock \href {https://doi.org/10.1109/CVPR.2019.01275} {Good news, everyone! context driven entity-aware captioning for news images}.
\newblock In \emph{2019 IEEE/CVF Conference on Computer Vision and Pattern Recognition (CVPR)}, pages 12458--12467.

\bibitem[{Caffagni et~al.(2024)Caffagni, Cocchi, Barsellotti, Moratelli, Sarto, Baraldi, Baraldi, Cornia, and Cucchiara}]{caffagni-etal-2024-revolution}
Davide Caffagni, Federico Cocchi, Luca Barsellotti, Nicholas Moratelli, Sara Sarto, Lorenzo Baraldi, Lorenzo Baraldi, Marcella Cornia, and Rita Cucchiara. 2024.
\newblock \href {https://doi.org/10.18653/v1/2024.findings-acl.807} {The revolution of multimodal large language models: A survey}.
\newblock In \emph{Findings of the Association for Computational Linguistics: ACL 2024}, pages 13590--13618, Bangkok, Thailand. Association for Computational Linguistics.

\bibitem[{Chen et~al.(2021)Chen, Huang, Lin, and Li}]{10.1145/3444685.3446322}
Aozhu Chen, Xinyi Huang, Hailan Lin, and Xirong Li. 2021.
\newblock \href {https://doi.org/10.1145/3444685.3446322} {Towards annotation-free evaluation of cross-lingual image captioning}.
\newblock In \emph{Proceedings of the 2nd ACM International Conference on Multimedia in Asia}, MMAsia '20, New York, NY, USA. Association for Computing Machinery.

\bibitem[{Costa-Juss{\`a} et~al.(2022)Costa-Juss{\`a}, Cross, Çelebi, Elbayad, Heafield, Heffernan, Kalbassi, Lam, Licht, Maillard, Sun, Wang, Wenzek, Youngblood, Akula, Barrault, Gonzalez, Hansanti, Hoffman, Jarrett, Sadagopan, Rowe, Spruit, Tran, Andrews, Ayan, Bhosale, Edunov, Fan, Gao, Goswami, Guzmán, Koehn, Mourachko, Ropers, Saleem, Schwenk, and Wang}]{costa2022no}
Marta~R Costa-Juss{\`a}, James~Cross Cross, Onur Çelebi, Maha Elbayad, Kenneth Heafield, Kevin Heffernan, Elahe Kalbassi, Janice Lam, Daniel Licht, Jean Maillard, Anna Sun, Skyler Wang, Guillaume Wenzek, Al~Youngblood, Bapi Akula, Loic Barrault, Gabriel~Mejia Gonzalez, Prangthip Hansanti, John Hoffman, Semarley Jarrett, Kaushik~Ram Sadagopan, Dirk Rowe, Shannon Spruit, Chau Tran, Pierre Andrews, Necip~Fazil Ayan, Shruti Bhosale, Sergey Edunov, Angela Fan, Cynthia Gao, Vedanuj Goswami, Francisco Guzmán, Philipp Koehn, Alexandre Mourachko, Christophe Ropers, Safiyyah Saleem, Holger Schwenk, and Jeff Wang. 2022.
\newblock No language left behind: Scaling human-centered machine translation.
\newblock \emph{arXiv preprint arXiv:2207.04672}.

\bibitem[{Daneshfar et~al.(2024)Daneshfar, Bartani, and Lotfi}]{DANESHFAR2024109288}
Fatemeh Daneshfar, Ako Bartani, and Pardis Lotfi. 2024.
\newblock \href {https://doi.org/https://doi.org/10.1016/j.engappai.2024.109288} {Image captioning by diffusion models: A survey}.
\newblock \emph{Engineering Applications of Artificial Intelligence}, 138:109288.

\bibitem[{Daud et~al.(2017)Daud, Khan, and Che}]{Daud2017}
Ali Daud, Wahab Khan, and Dunren Che. 2017.
\newblock \href {https://doi.org/10.1007/s10462-016-9482-x} {Urdu language processing: a survey}.
\newblock \emph{Artificial Intelligence Review}, 47(3):279--311.

\bibitem[{De~Silva(2019)}]{de2019survey}
Nisansa De~Silva. 2019.
\newblock Survey on publicly available sinhala natural language processing tools and research.
\newblock \emph{arXiv preprint arXiv:1906.02358}.

\bibitem[{Farabi et~al.(2024)Farabi, Ranasinghe, Kanojia, Kong, and Zampieri}]{10.24963/ijcai.2024/887}
Shafkat Farabi, Tharindu Ranasinghe, Diptesh Kanojia, Yu~Kong, and Marcos Zampieri. 2024.
\newblock \href {https://doi.org/10.24963/ijcai.2024/887} {A survey of multimodal sarcasm detection}.
\newblock In \emph{Proceedings of the Thirty-Third International Joint Conference on Artificial Intelligence}, IJCAI '24.

\bibitem[{Ghandi et~al.(2023)Ghandi, Pourreza, and Mahyar}]{10.1145/3617592}
Taraneh Ghandi, Hamidreza Pourreza, and Hamidreza Mahyar. 2023.
\newblock \href {https://doi.org/10.1145/3617592} {Deep learning approaches on image captioning: A review}.
\newblock \emph{ACM Comput. Surv.}, 56(3).

\bibitem[{Hasan et~al.(2021)Hasan, Bhattacharjee, Islam, Mubasshir, Li, Kang, Rahman, and Shahriyar}]{hasan-etal-2021-xl}
Tahmid Hasan, Abhik Bhattacharjee, Md.~Saiful Islam, Kazi Mubasshir, Yuan-Fang Li, Yong-Bin Kang, M.~Sohel Rahman, and Rifat Shahriyar. 2021.
\newblock \href {https://doi.org/10.18653/v1/2021.findings-acl.413} {{XL}-sum: Large-scale multilingual abstractive summarization for 44 languages}.
\newblock In \emph{Findings of the Association for Computational Linguistics: ACL-IJCNLP 2021}, pages 4693--4703, Online. Association for Computational Linguistics.

\bibitem[{Hettiarachchi et~al.(2024)Hettiarachchi, Premasiri, Uyangodage, and Ranasinghe}]{hettiarachchi-etal-2024-nsina}
Hansi Hettiarachchi, Damith Premasiri, Lasitha Randunu~Chandrakantha Uyangodage, and Tharindu Ranasinghe. 2024.
\newblock \href {https://aclanthology.org/2024.lrec-main.1076/} {{NS}ina: A news corpus for {S}inhala}.
\newblock In \emph{Proceedings of the 2024 Joint International Conference on Computational Linguistics, Language Resources and Evaluation (LREC-COLING 2024)}, pages 12307--12312, Torino, Italia. ELRA and ICCL.

\bibitem[{Joshi et~al.(2020)Joshi, Santy, Budhiraja, Bali, and Choudhury}]{joshi-etal-2020-state}
Pratik Joshi, Sebastin Santy, Amar Budhiraja, Kalika Bali, and Monojit Choudhury. 2020.
\newblock \href {https://doi.org/10.18653/v1/2020.acl-main.560} {The state and fate of linguistic diversity and inclusion in the {NLP} world}.
\newblock In \emph{Proceedings of the 58th Annual Meeting of the Association for Computational Linguistics}, pages 6282--6293, Online. Association for Computational Linguistics.

\bibitem[{Kim et~al.(2023)Kim, Hwang, Yun, Yoon, Bui, and Jung}]{kim-etal-2023-pr}
Yongil Kim, Yerin Hwang, Hyeongu Yun, Seunghyun Yoon, Trung Bui, and Kyomin Jung. 2023.
\newblock \href {https://doi.org/10.18653/v1/2023.findings-emnlp.819} {{PR}-{MCS}: Perturbation robust metric for {M}ulti{L}ingual image captioning}.
\newblock In \emph{Findings of the Association for Computational Linguistics: EMNLP 2023}, pages 12237--12258, Singapore. Association for Computational Linguistics.

\bibitem[{Li et~al.(2024)Li, Ge, Ge, Wang, Wang, Zhang, and Shan}]{li2024seed}
Bohao Li, Yuying Ge, Yixiao Ge, Guangzhi Wang, Rui Wang, Ruimao Zhang, and Ying Shan. 2024.
\newblock Seed-bench: Benchmarking multimodal large language models.
\newblock In \emph{Proceedings of the IEEE/CVF Conference on Computer Vision and Pattern Recognition}, pages 13299--13308.

\bibitem[{Li et~al.(2022)Li, Li, Xiong, and Hoi}]{pmlr-v162-li22n}
Junnan Li, Dongxu Li, Caiming Xiong, and Steven Hoi. 2022.
\newblock \href {https://proceedings.mlr.press/v162/li22n.html} {{BLIP}: Bootstrapping language-image pre-training for unified vision-language understanding and generation}.
\newblock In \emph{Proceedings of the 39th International Conference on Machine Learning}, volume 162 of \emph{Proceedings of Machine Learning Research}, pages 12888--12900. PMLR.

\bibitem[{Liang et~al.(2024)Liang, Xu, Hong, Shang, Wang, Fu, and Liu}]{10.1145/3672758.3672824}
Zijing Liang, Yanjie Xu, Yifan Hong, Penghui Shang, Qi~Wang, Qiang Fu, and Ke~Liu. 2024.
\newblock \href {https://doi.org/10.1145/3672758.3672824} {A survey of multimodel large language models}.
\newblock In \emph{Proceedings of the 3rd International Conference on Computer, Artificial Intelligence and Control Engineering}, CAICE '24, page 405–409, New York, NY, USA. Association for Computing Machinery.

\bibitem[{Lin et~al.(2014)Lin, Maire, Belongie, Hays, Perona, Ramanan, Doll{\'a}r, and Zitnick}]{10.1007/978-3-319-10602-1_48}
Tsung-Yi Lin, Michael Maire, Serge Belongie, James Hays, Pietro Perona, Deva Ramanan, Piotr Doll{\'a}r, and C.~Lawrence Zitnick. 2014.
\newblock Microsoft coco: Common objects in context.
\newblock In \emph{Computer Vision -- ECCV 2014}, pages 740--755, Cham. Springer International Publishing.

\bibitem[{Liu et~al.(2021)Liu, Wang, Wang, and Ordonez}]{liu-etal-2021-visual}
Fuxiao Liu, Yinghan Wang, Tianlu Wang, and Vicente Ordonez. 2021.
\newblock \href {https://doi.org/10.18653/v1/2021.emnlp-main.542} {Visual news: Benchmark and challenges in news image captioning}.
\newblock In \emph{Proceedings of the 2021 Conference on Empirical Methods in Natural Language Processing}, pages 6761--6771, Online and Punta Cana, Dominican Republic. Association for Computational Linguistics.

\bibitem[{Mathur et~al.(2024)Mathur, Liu, Li, Ma, Karen, Ahmed, Manocha, and Zhang}]{mathur-etal-2024-doc}
Puneet Mathur, Zhe Liu, Ke~Li, Yingyi Ma, Gil Karen, Zeeshan Ahmed, Dinesh Manocha, and Xuedong Zhang. 2024.
\newblock \href {https://aclanthology.org/2024.lrec-main.457/} {{DOC}-{RAG}: {ASR} language model personalization with domain-distributed co-occurrence retrieval augmentation}.
\newblock In \emph{Proceedings of the 2024 Joint International Conference on Computational Linguistics, Language Resources and Evaluation (LREC-COLING 2024)}, pages 5132--5139, Torino, Italia. ELRA and ICCL.

\bibitem[{Nussbaum et~al.(2024)Nussbaum, Duderstadt, and Mulyar}]{nussbaum2024nomic}
Zach Nussbaum, Brandon Duderstadt, and Andriy Mulyar. 2024.
\newblock Nomic embed vision: Expanding the latent space.
\newblock \emph{arXiv preprint arXiv:2406.18587}.

\bibitem[{Oliveira~dos Santos et~al.(2021)Oliveira~dos Santos, Colombini, and Avila}]{oliveira-dos-santos-etal-2021-cider}
Gabriel Oliveira~dos Santos, Esther~Luna Colombini, and Sandra Avila. 2021.
\newblock \href {https://doi.org/10.18653/v1/2021.wnut-1.39} {{CIDE}r-{R}: Robust consensus-based image description evaluation}.
\newblock In \emph{Proceedings of the Seventh Workshop on Noisy User-generated Text (W-NUT 2021)}, pages 351--360, Online. Association for Computational Linguistics.

\bibitem[{Papineni et~al.(2002)Papineni, Roukos, Ward, and Zhu}]{papineni-etal-2002-bleu}
Kishore Papineni, Salim Roukos, Todd Ward, and Wei-Jing Zhu. 2002.
\newblock \href {https://doi.org/10.3115/1073083.1073135} {{B}leu: a method for automatic evaluation of machine translation}.
\newblock In \emph{Proceedings of the 40th Annual Meeting of the Association for Computational Linguistics}, pages 311--318, Philadelphia, Pennsylvania, USA. Association for Computational Linguistics.

\bibitem[{Qu et~al.(2024)Qu, Tuytelaars, and Moens}]{qu-etal-2024-visually}
Tingyu Qu, Tinne Tuytelaars, and Marie-Francine Moens. 2024.
\newblock \href {https://doi.org/10.18653/v1/2024.naacl-long.162} {Visually-aware context modeling for news image captioning}.
\newblock In \emph{Proceedings of the 2024 Conference of the North American Chapter of the Association for Computational Linguistics: Human Language Technologies (Volume 1: Long Papers)}, pages 2927--2943, Mexico City, Mexico. Association for Computational Linguistics.

\bibitem[{Rajakumar~Kalarani et~al.(2023)Rajakumar~Kalarani, Bhattacharyya, Chhaya, and Shekhar}]{rajakumar-kalarani-etal-2023-lets}
Abisek Rajakumar~Kalarani, Pushpak Bhattacharyya, Niyati Chhaya, and Sumit Shekhar. 2023.
\newblock \href {https://doi.org/10.18653/v1/2023.acl-industry.67} {{\textquotedblleft}let`s not quote out of context{\textquotedblright}: Unified vision-language pretraining for context assisted image captioning}.
\newblock In \emph{Proceedings of the 61st Annual Meeting of the Association for Computational Linguistics (Volume 5: Industry Track)}, pages 695--706, Toronto, Canada. Association for Computational Linguistics.

\bibitem[{Ramos et~al.(2023)Ramos, Martins, and Elliott}]{ramos-etal-2023-lmcap}
Rita Ramos, Bruno Martins, and Desmond Elliott. 2023.
\newblock \href {https://doi.org/10.18653/v1/2023.findings-acl.104} {{LMC}ap: Few-shot multilingual image captioning by retrieval augmented language model prompting}.
\newblock In \emph{Findings of the Association for Computational Linguistics: ACL 2023}, pages 1635--1651, Toronto, Canada. Association for Computational Linguistics.

\bibitem[{Ranasinghe et~al.(2025{\natexlab{a}})Ranasinghe, Hettiarachchi, Orasan, and Mitkov}]{ranasinghe-etal-2025-musts}
Tharindu Ranasinghe, Hansi Hettiarachchi, Constantin Orasan, and Ruslan Mitkov. 2025{\natexlab{a}}.
\newblock \href {https://doi.org/10.18653/v1/2025.acl-short.27} {{MUSTS}: {MU}ltilingual semantic textual similarity benchmark}.
\newblock In \emph{Proceedings of the 63rd Annual Meeting of the Association for Computational Linguistics (Volume 2: Short Papers)}, pages 331--353, Vienna, Austria. Association for Computational Linguistics.

\bibitem[{Ranasinghe et~al.(2025{\natexlab{b}})Ranasinghe, Hettiarachchi, Pathirana, Premasiri, Uyangodage, Nanomi~Arachchige, Plum, Rayson, and Mitkov}]{ranasinghe-etal-2025-sinhala}
Tharindu Ranasinghe, Hansi Hettiarachchi, Nadeesha Chathurangi Naradde~Vidana Pathirana, Damith Premasiri, Lasitha Uyangodage, Isuri Nanomi~Arachchige, Alistair Plum, Paul Rayson, and Ruslan Mitkov. 2025{\natexlab{b}}.
\newblock \href {https://doi.org/10.18653/v1/2025.acl-long.422} {{S}inhala encoder-only language models and evaluation}.
\newblock In \emph{Proceedings of the 63rd Annual Meeting of the Association for Computational Linguistics (Volume 1: Long Papers)}, pages 8623--8636, Vienna, Austria. Association for Computational Linguistics.

\bibitem[{Sellam et~al.(2020)Sellam, Das, and Parikh}]{sellam-etal-2020-bleurt}
Thibault Sellam, Dipanjan Das, and Ankur Parikh. 2020.
\newblock \href {https://doi.org/10.18653/v1/2020.acl-main.704} {{BLEURT}: Learning robust metrics for text generation}.
\newblock In \emph{Proceedings of the 58th Annual Meeting of the Association for Computational Linguistics}, pages 7881--7892, Online. Association for Computational Linguistics.

\bibitem[{Stefanini et~al.(2023)Stefanini, Cornia, Baraldi, Cascianelli, Fiameni, and Cucchiara}]{9706348}
Matteo Stefanini, Marcella Cornia, Lorenzo Baraldi, Silvia Cascianelli, Giuseppe Fiameni, and Rita Cucchiara. 2023.
\newblock \href {https://doi.org/10.1109/TPAMI.2022.3148210} {{ From Show to Tell: A Survey on Deep Learning-Based Image Captioning }}.
\newblock \emph{IEEE Transactions on Pattern Analysis \& Machine Intelligence}, 45(01):539--559.

\bibitem[{Tran et~al.(2020)Tran, Mathews, and Xie}]{9156456}
Alasdair Tran, Alexander Mathews, and Lexing Xie. 2020.
\newblock \href {https://doi.org/10.1109/CVPR42600.2020.01305} {Transform and tell: Entity-aware news image captioning}.
\newblock In \emph{2020 IEEE/CVF Conference on Computer Vision and Pattern Recognition (CVPR)}, pages 13032--13042.

\bibitem[{Wu et~al.(2023{\natexlab{a}})Wu, Gan, Chen, Wan, and Yu}]{10386743}
Jiayang Wu, Wensheng Gan, Zefeng Chen, Shicheng Wan, and Philip~S. Yu. 2023{\natexlab{a}}.
\newblock \href {https://doi.org/10.1109/BigData59044.2023.10386743} {Multimodal large language models: A survey}.
\newblock In \emph{2023 IEEE International Conference on Big Data (BigData)}, pages 2247--2256.

\bibitem[{Wu et~al.(2023{\natexlab{b}})Wu, Fei, Ji, and Chua}]{wu-etal-2023-cross2stra}
Shengqiong Wu, Hao Fei, Wei Ji, and Tat-Seng Chua. 2023{\natexlab{b}}.
\newblock \href {https://doi.org/10.18653/v1/2023.acl-long.146} {{C}ross2{S}tr{A}: Unpaired cross-lingual image captioning with cross-lingual cross-modal structure-pivoted alignment}.
\newblock In \emph{Proceedings of the 61st Annual Meeting of the Association for Computational Linguistics (Volume 1: Long Papers)}, pages 2593--2608, Toronto, Canada. Association for Computational Linguistics.

\bibitem[{Xia et~al.(2024)Xia, Zhu, Li, Zhu, Li, Li, Zhang, and Yao}]{xia-etal-2024-rule}
Peng Xia, Kangyu Zhu, Haoran Li, Hongtu Zhu, Yun Li, Gang Li, Linjun Zhang, and Huaxiu Yao. 2024.
\newblock \href {https://doi.org/10.18653/v1/2024.emnlp-main.62} {{RULE}: Reliable multimodal {RAG} for factuality in medical vision language models}.
\newblock In \emph{Proceedings of the 2024 Conference on Empirical Methods in Natural Language Processing}, pages 1081--1093, Miami, Florida, USA. Association for Computational Linguistics.

\bibitem[{Xiao et~al.(2025)Xiao, Zhou, Liu, Liu, Li, Liu, and Huang}]{XIAO2025102888}
Hanguang Xiao, Feizhong Zhou, Xingyue Liu, Tianqi Liu, Zhipeng Li, Xin Liu, and Xiaoxuan Huang. 2025.
\newblock \href {https://doi.org/https://doi.org/10.1016/j.inffus.2024.102888} {A comprehensive survey of large language models and multimodal large language models in medicine}.
\newblock \emph{Information Fusion}, 117:102888.

\bibitem[{Xu et~al.(2023{\natexlab{a}})Xu, Tang, Lv, Zheng, Zeng, and Li}]{XU2023126287}
Liming Xu, Quan Tang, Jiancheng Lv, Bochuan Zheng, Xianhua Zeng, and Weisheng Li. 2023{\natexlab{a}}.
\newblock \href {https://doi.org/https://doi.org/10.1016/j.neucom.2023.126287} {Deep image captioning: A review of methods, trends and future challenges}.
\newblock \emph{Neurocomputing}, 546:126287.

\bibitem[{Xu et~al.(2023{\natexlab{b}})Xu, Hu, Zhou, Hao, and Hong}]{10.1145/3582649.3582658}
Yueyuan Xu, Zhenzhen Hu, Yuanen Zhou, Shijie Hao, and Richang Hong. 2023{\natexlab{b}}.
\newblock \href {https://doi.org/10.1145/3582649.3582658} {Cite: Compact interactive transformer for multilingual image captioning}.
\newblock In \emph{Proceedings of the 2023 6th International Conference on Image and Graphics Processing}, ICIGP '23, page 175–181, New York, NY, USA. Association for Computing Machinery.

\bibitem[{Young et~al.(2014)Young, Lai, Hodosh, and Hockenmaier}]{young-etal-2014-image}
Peter Young, Alice Lai, Micah Hodosh, and Julia Hockenmaier. 2014.
\newblock \href {https://doi.org/10.1162/tacl_a_00166} {From image descriptions to visual denotations: New similarity metrics for semantic inference over event descriptions}.
\newblock \emph{Transactions of the Association for Computational Linguistics}, 2:67--78.

\bibitem[{Zhang et~al.(2024)Zhang, Guo, Yang, Song, and Wang}]{10.1145/3634917}
Jing Zhang, Dan Guo, Xun Yang, Peipei Song, and Meng Wang. 2024.
\newblock \href {https://doi.org/10.1145/3634917} {Visual-linguistic-stylistic triple reward for cross-lingual image captioning}.
\newblock \emph{ACM Trans. Multimedia Comput. Commun. Appl.}, 20(4).

\bibitem[{Zhang* et~al.(2020)Zhang*, Kishore*, Wu*, Weinberger, and Artzi}]{Zhang*2020BERTScore}
Tianyi Zhang*, Varsha Kishore*, Felix Wu*, Kilian~Q. Weinberger, and Yoav Artzi. 2020.
\newblock \href {https://openreview.net/forum?id=SkeHuCVFDr} {Bertscore: Evaluating text generation with bert}.
\newblock In \emph{International Conference on Learning Representations}.

\end{thebibliography}
